\documentclass[runningheads]{llncs}
\usepackage{booktabs}
\usepackage{xcolor}
\usepackage{mathtools}
\usepackage{amsmath,amssymb,amsfonts}
\usepackage{graphicx}
\usepackage[final]{listings}
\usepackage{subcaption}

\DeclarePairedDelimiterX{\norm}[1]{\lVert}{\rVert}{#1}

\newcommand{\argmin}[1]{\underset{#1}{\operatorname{arg}\,\operatorname{min}}\;}
\newcommand{\argmax}[1]{\underset{#1}{\operatorname{arg}\,\operatorname{max}}\;}

\usepackage{adjustbox}
\usepackage{array}
\usepackage{hyperref}

\makeatletter
\newcommand{\printfnsymbol}[1]{%
  \textsuperscript{\@fnsymbol{#1}}%
}
\makeatother

\begin{document}

\title{Benchmarking adversarial attacks and defenses for time-series data}
%
%
\author{Shoaib Ahmed Siddiqui\inst{1}\inst{2}\orcidID{0000-0003-4600-7331} \and Andreas Dengel\inst{1}\inst{2}\orcidID{0000-0002-6100-8255} \and Sheraz Ahmed\inst{1}\orcidID{0000-0002-4239-6520}}
\authorrunning{S. A. Siddiqui et al.}
%
\institute{German Research Center for Artificial Intelligence (DFKI) \and TU Kaiserslautern \\
\email{first_name.last_name@dfki.de}}
\institute{German Research Center for Artificial Intelligence (DFKI), 67663 Kaiserslautern, Germany \\ 
\email{first\_name.last\_name@dfki.de} \and
TU Kaiserslautern, 67663 Kaiserslautern, Germany}
\maketitle              

\begin{abstract}
The adversarial vulnerability of deep networks has spurred the interest of researchers worldwide. Unsurprisingly, like images, adversarial examples also translate to time-series data as they are an inherent weakness of the model itself rather than the modality. Several attempts have been made to defend against these adversarial attacks, particularly for the visual modality. In this paper, we perform detailed benchmarking of well-proven adversarial defense methodologies on time-series data. We restrict ourselves to the $L_{\infty}$ threat model. We also explore the trade-off between smoothness and clean accuracy for regularization-based defenses to better understand the trade-offs that they offer. Our analysis shows that the explored adversarial defenses offer robustness against both strong white-box as well as black-box attacks. This paves the way for future research in the direction of adversarial attacks and defenses, particularly for time-series data.
\end{abstract}

\section{Introduction}

Time-series data is ubiquitous in this era of internet-of-things (IoT) and industry 4.0 where millions of sensors are generating data at an extremely high frequency~\cite{siddiqui2019tsviz,fawaz2019adversarial,karim2020adversarial,harford2020adversarial}. With this increasing amount of data, there has been a wide-scale deployment of deep models for time-series analysis. 
Deep learning models have proven to be susceptible to changes in the input which can significantly alter the predictions of the classifier~\cite{szegedy2013intriguing}. This raises a serious concern over the real-world deployment of these models. This vulnerability has been particularly explored in the context of images~\cite{szegedy2013intriguing,goodfellow2014FGSM,madry2017towards,brendel2017decision}. 

A similar vulnerability exists also in the context of other modalities as this is primarily a weakness of the current learning paradigm rather than the modality~\cite{szegedy2013intriguing}. Since many of the time-series models are deployed security-critical scenarios, there has been an increasing interest regarding the robustness of these models~\cite{siddiqui2019tsviz,fawaz2019adversarial,karim2020adversarial,harford2020adversarial}. Therefore, gradient-based attacks prevalent in the computer-vision community have also been translated to time-series data~\cite{siddiqui2019tsviz,fawaz2019adversarial,karim2020adversarial}.

Various defenses have been proposed to circumvent this adversarial vulnerability~\cite{madry2017towards,zhang2019trades,xie2019featuredenoising}. Some of these defenses are specific to the visual modality, while a wide range of literature has been focused on either training on these adversarial examples or adding an additional regularization term that forces the predictions to be consistent within a specific neighborhood of the example.
In this paper, we employ some of the most well-recognized defense methodologies tested on images and evaluate their robustness for time-series data to establish a proper benchmark.

\section{Related Work}  \label{sec:related}

Adversarial examples were first discovered by Szegedy et al. (2013)~\cite{szegedy2013intriguing} as a consequence of trying to solve an inverse optimization problem. Since then, a wide range of literature has focused on this security aspect including both development of more sophisticated defenses as well as advances in attacks in order to break these defenses.

Adversarial attacks can be mainly categorized into two different categories namely white-box attacks and black-box attacks. White-box attacks assume access to model architecture and parameters, therefore, they can effectively and efficiently attack the model using the gradient information. Black-box attacks, on the other hand, require access to either the output probabilities or even just the label, making them more applicable in real-world settings. However, black-box attacks usually require thousands or even millions of queries to the model to compute just a single adversarial example.

Szegedy et al. (2013)~\cite{szegedy2013intriguing} presented the first adversarial attack based on box-constrained L-BFGS to mine adversarial examples. Goodfellow et al. (2014)~\cite{goodfellow2014FGSM} proposed a fast version of the attack by assuming the linearity of classifiers around the input. With this assumption, they were able to use a single step attack based on the gradient which they named Fast Gradient-Sign Method (FGSM). Madry et al. (2017)~\cite{madry2017towards} proposed an iterative version of FGSM with a random restart which they named Projected Gradient Descent (PGD) and claimed it to be an optimal first-order adversary. Another famous attack is Carlini-Wagner~\cite{carlini2017towards} attack. However, this is mainly designed for $L_2$ norm-based attacks while we only focus on $L_{\infty}$ norm-based attacks in this paper.
Boundary attack~\cite{brendel2017decision} introduced by Brendel et al. (2017) formed the basis for decision-based attacks where the attacker assumes access only to the output label. SIMple Black-box Attack (SIMBA)~\cite{simba} proposed by Guo et al. (2019) greatly simplified the attack pipeline by assuming access to the output probabilities of the model. The attack mines adversarial examples by just randomly perturbing pixels if they have a negative impact on the output probability. 

As attacks have progressed, more and more sophisticated methods have been developed to defend against these attacks. However, most of these attacks were either shown to be masking the gradient or poorly tested~\cite{obfuscated_gradients}. 
One of the most effective methods to defend against adversarial attacks is PGD-based adversarial training~\cite{madry2017towards}. The robust model is trained on the generated adversarial examples rather than the original inputs in this case. As PGD is one of the most powerful white-box attacks, PGD-based adversarially trained models were shown to be significantly superior in terms of robustness as compared to other techniques~\cite{madry2017towards,obfuscated_gradients}. 
Zhang et al. (2019)~\cite{zhang2019trades} proposed TRADES which uses an additional regularization term along with the conventional cross-entropy loss that minimizes the discrepancy between clean and adversarial predictions. 
Feature denoising~\cite{xie2019featuredenoising} proposed by Zie et al. (2019) introduced additional denoising operators in the network. The whole network was then trained using adversarial training~\cite{madry2017towards}. The idea was based on minimizing the discrepancy between the feature maps of a clean and adversarial example.
A large fraction of adversarial defense literature has been focused on provable defenses that provide formal guarantees against the worst-case adversary. However, they are usually prohibitively slow and unable to scale to large datasets~\cite{wong2018provable}. We don't include these methods in our comparison and leave it as future work.

Minor efforts have also been made in terms of extending these attacks for time-series data. Siddiqui et al. (2019)~\cite{siddiqui2019tsviz} showed that gradient-based adversarial attacks were effective for both time-series classification as well as time-series regression networks. Fawaz et al. (2019)~\cite{fawaz2019adversarial} analyzed a range of different time-series datasets and showed that deep models trained on time-series data are vulnerable to adversarial attacks. Both these papers only explore the vulnerability of networks, which is not surprising given that adversarial attacks exploit the machine learning optimization framework, rather than a specific modality~\cite{goodfellow2014FGSM}. Karim et al. (2020)~\cite{karim2020adversarial} and Harford et al. (2020)~\cite{harford2020adversarial} employed Gradient Adversarial Transformation Network (GATN) for attacking models. Since they also considered classical time-series models that are non-differentiable, they used a knowledge distillation approach to train a student network mimicking the predictions of the original classifier. Therefore, what they explored were just transfer attacks which are a rather weak form of black-box attacks. On the other hand, we evaluate using both the strongest white-box as well as black-box attacks to truly establish the robustness of the evaluated defenses.
Our work specifies a proper threat model when evaluating against attacks as well as considers both strong white-box and black-box attacks which haven't been explored in the context of time-series data. Therefore, we not only benchmark adversarial defenses but also establish a benchmark for strong adversarial attacks to be considered for future work which has been missing in the prior work~\cite{carlini2019evaluating}.

\section{Method} \label{sec:method}

\subsection{Threat Model}

The threat model specifies the conditions under which the considered defense is designed to be secure~\cite{carlini2019evaluating}. 
We consider $L_{\infty}$ threat model and use an epsilon of $0.3$ for training robust models. There are no box-constraints for time-series data due to their variable input range in contrast to the visual modality where the pixels take on a discrete value from $[0, 255]$. Therefore, in our case, the data was normalized with zero-mean and unit standard deviation which justified the choice of 0.3 as the epsilon value.

\subsection{Adversarially Robust Models (using Adversarial Defenses)} \label{sec:defenses}

A robust model is a model which is robust against these minor corruptions of the input signal. A robust model is usually obtained by training a model with a particular adversarial defense methodology. There is a very wide range of literature on the topic of adversarial defenses. However, most of these defenses were shown to be broken by a stronger attack~\cite{obfuscated_gradients}. Therefore, we only explored defenses that withstood these attacks when evaluated on images. We will now discuss each of the evaluated defenses in detail.

\subsubsection{Adversarial training}

Madry et al. (2017)~\cite{madry2017towards} proposed a robust optimization algorithm which they named as adversarial training. Adversarial training is one of the most simple and widely accepted adversarial defenses in the literature. The idea is to just train a classifier on the attacked examples rather than the original ones. As the model inherently learns to be robust to these attacks during training, this naturally re-configures the decision boundaries of the network. 

\begin{equation*}
    \mathbf{w}^{*} = \argmin{\mathbf{w}} \;\; \argmax{\mathbf{x'} \in B_p(\mathbf{x}, \epsilon)} \mathcal{L}(\Phi(\mathbf{x'}; \mathbf{w}), y')
\end{equation*}

\noindent where $y'$ indicates the model's prediction on the input $\mathbf{x}$. The maximization problem is approximately solved by generating an adversarial example using the PGD attack which is considered to be the optimal first-order adversary~\cite{madry2017towards}. The biggest advantage of adversarial training is that there are no additional hyperparameters making model training very easy and convenient.

\subsubsection{TRADES}

Zhang et al. (2019)~\cite{zhang2019trades} introduced TRADES which smoothens the predictions of the classifier around the input by employing an additional term in the final objective alongside the conventional cross-entropy.

\begin{equation*}
    \mathbf{w}^{*} = \argmin{\mathbf{w}} \;\; \Big\{ \mathcal{L}_{CE}(\Phi(\mathbf{x}; \mathbf{w}), y) + \argmax{\mathbf{x'} \in B_p(\mathbf{x}, \epsilon)} \mathcal{L}_{KL}(\Phi(\mathbf{x}; \mathbf{w}), \Phi(\mathbf{x'}; \mathbf{w})) / \lambda \Big\}
\end{equation*}

\noindent where $\mathcal{L}_{CE}$ represents the conventional cross-entropy loss on clean data while $\mathcal{L}_{KL}$ computes the KL-divergence between the logits obtained from the original example and the computed adversarial example. The maximization problem is again approximately solved by generating an adversarial example using the PGD attack. TRADES introduces an additional hyperparameter $\lambda$ which controls the trade-off between clean accuracy and adversarial robustness.

\subsubsection{Feature Denoising}

Xie et al. (2019)~\cite{xie2019featuredenoising} introduced the idea of feature denoising based on their observation that the feature maps computed for an adversarial example are significantly noiser than the ones computed for the original image. Therefore, in order to circumvent this problem, they used denoising operators before max-pooling layers in their network. They compared different denoising operators, and found Gaussian Non-Local Means (GNLM) to be the most effective one. This can be represented as:

\begin{equation*}
    y_i = \frac{1}{\sum_{\forall j \in \mathcal{N}} f(x_i, x_j)} \sum_{\forall j \in \mathcal{N}} f(x_i, x_j) \times x_j
\end{equation*}

\noindent where $y_i$ denotes the $i^{th}$ output, $\mathcal{N}$ denotes all the spatial locations on the feature map and $f(x_i, x_j)$ captures the similarity between $x_i$ and $x_j$. Since we use the Gaussian version of non-local means, the similarity function is given by $f(x_i, x_j)=e^{\frac{1}{\sqrt{d}} \theta(x_i)^t \phi(x_j)}$ where $\theta(x_i) \in \mathbb{R}^{64}$ and $\phi(x_j) \in \mathbb{R}^{64}$ represents two embedded versions of the input implemented via $1 \times 1$ convolution, and $d$ denotes the number of channels. 
Based on their findings, we also introduce an additional GLNM denoising layer before every pooling layer in all of our networks. Since this denoising layer is also learned during adversarial training, we compare it's impact when training the model using different adversarial defense techniques.



\subsection{Robust Evaluation (using Adversarial Attacks)} \label{sec:attacks}

The aim of robust evaluation is to precisely identify the robustness of the trained robust model. Carlini et al.~\cite{carlini2019evaluating} provided comprehensive guidelines to evaluate robust models in order to avoid pitfalls which prior defense methods could not avoid, providing a false sense of security.
For this reason, we included two major black-box as well as two major white-box attacks to compare. In total, we used 5 different attacks, where one black-box attack i.e. noise attack is a rather weak attack, but serves as a trivial baseline.

\noindent \underline{\textbf{Evaluation Metric}}

\noindent There are many different choices when evaluating robust models. This includes the input examples to consider as well as the target to use for computing the adversarial example. In our case, we compute robust accuracy on all examples regardless of whether they were correctly classified or not. We always conduct attack using the original labels instead of the model's prediction in order to ensure that the attack does not mistakenly move an incorrectly classified example to a correct class, providing a false sense of robustness. 
However, regardless of the choice of this metric, the key takeaways from our experiments still remain the same.
We evaluate robustness using untargeted attacks as they can be considered worst-case adversaries. Targeted attacks are usually much more relevant in practice but harder to find as compared to untargeted attacks.

\noindent \underline{\textbf{White-box Attacks}}

\noindent \textbf{FGSM}~\cite{goodfellow2014FGSM}: Goodfellow et al. (2014)~\cite{goodfellow2014FGSM} posited that the lack of robustness of deep models is due to their linear nature. Therefore, they used this linear approximation to develop a Fast Gradient-Sign based attack Method (FGSM) which takes just a single step in the direction pointed by the gradient.

\begin{equation*}
    \mathbf{x}_{a} = \mathbf{x} + \epsilon \; sign \big( \frac{\partial}{\partial \mathbf{x}} \mathcal{L}(\Phi(\mathbf{x}; \mathbf{w}), y) \big)
\end{equation*}

\noindent where $y$ can either be the original label or the model's prediction, and $sign$ returns the sign of the gradient. We used 100 random restarts for FGSM-100 and picked the best adversarial example from these 100 restarts.

\noindent \textbf{PGD}~\cite{madry2017towards}: Projected-Gradient Descent (PGD) is an iterative version of FGSM with an additional random restart.

\begin{equation*}
    \mathbf{x}_a^{(t+1)} = Clip_{\mathbf{x}, \epsilon} \Big\{ \mathbf{x}_a^{(t)} + \alpha \; sign \big( \frac{\partial}{\partial \mathbf{x}_a^{(t)}} \mathcal{L}(\Phi(\mathbf{x}_a^{(t)}; \mathbf{w}), y) \big) \Big\}
\end{equation*}

\noindent where $\alpha$ is per-step update size, $Clip_{\mathbf{x}, \epsilon}$ binds the $L_{\infty}$ norm of the perturbation to be $\epsilon$, and $\mathbf{x}_a^{(t)}$ indicates the adversarial example obtained after the $t^{th}$ optimization step. In the case of PGD, $\mathbf{x}_a^{(0)} = \mathbf{x} + \delta$ where $\delta$ is a random perturbation within the $L_{\infty}$ norm-ball. We used 10 random restarts for PGD-10 and picked the best adversarial example from these 10 restarts. Each every restart, we use 100 PGD steps with $\alpha=2 \times \epsilon/T$ where $T$ is the number of PGD steps.


\noindent \underline{\textbf{Black-box Attacks}}

\noindent \textbf{Noise Attack}: One of the most simple and preliminary attacks is the random noise attack. The idea is to generate a set of random vectors and pick the best one out of them. We used a set of 100 random vectors for NOISE-100.



\noindent \textbf{Boundary Attack}~\cite{brendel2017decision}: Boundary attack was the first decision-based attack. It starts with a random input that is not classified as the given label $y$ and walks towards the original input until it hits the decision boundary. At this point, the attack starts moving orthogonal to the decision boundary until it encounters the closest attainable point to the given example $\mathbf{x}$. Boundary attack minimizes the $L_2$ norm of perturbation, therefore, it is not particularly optimized for the $L_{\infty}$ models that we consider in our case.
In order to compute the robust accuracy metric that we report, we compute the $L_{\infty}$ norm of the computed perturbation and then discard any perturbations which exceed this budget. Therefore, in many cases, although being a powerful attack when considering euclidean distances, it does not provide high success rates when evaluated on $L_{\infty}$ norm.

\noindent \textbf{Simple Black-box Attack (SIMBA)}~\cite{simba}: SIMBA is one of the most simple black-box attacks which uses the predicted probabilities from the classifier to decide the perturbation vector. SIMBA attack considers all of the input points in the sequence and computes their impact on the probability of the predicted class if perturbed by either $\{-\epsilon, \epsilon\}$. It then chooses perturbation, whichever maximally reduces the probability of the predicted class. If the probability of the predicted class is not impacted by the chosen input point, the attack leaves it intact, hence, it also minimizes the $L_0$ norm of the perturbation. The number of input points queried to compute the adversarial example is restricted when considering high-dimensional images in order to make the attack time efficient. However, we did not impose any such restrictions as the number of points in a sequence is usually smaller in contrast to the number of pixels in an image.

\subsection{Dataset} \label{sec:dataset}

Due to lack of space, we only present results on the famous character trajectories dataset\footnote{\url{https://archive.ics.uci.edu/ml/datasets/Character+Trajectories}}. We also experimented with several other datasets and found the results to be consistent between these different datasets. The character trajectories dataset contains hand-written characters using a Wacom tablet. Only three dimensions are kept for the final dataset which includes x, y and pen-tip force. 
The sampling rate was set to be 200 Hz. The data was numerically differentiated and Gaussian smoothen with $\sigma = 2$. The task is to classify the characters into 20 different classes.
This dataset is comprised of 2858 character samples divided into 1383 training, 606 validation and 869 test sequences. Each sequence is comprised of 206 time-steps with three channels. 
Since we need to constrain the input range for precisely defining the epsilon norm-ball to consider within our attack and defense framework, we normalize the data to have zero mean and unit standard deviation. 

\section{Results}  \label{sec:results}

    
    

\begin{figure}[t]
    \centering
    \begin{subfigure}[b]{0.31\textwidth}
        \includegraphics[width=\textwidth]{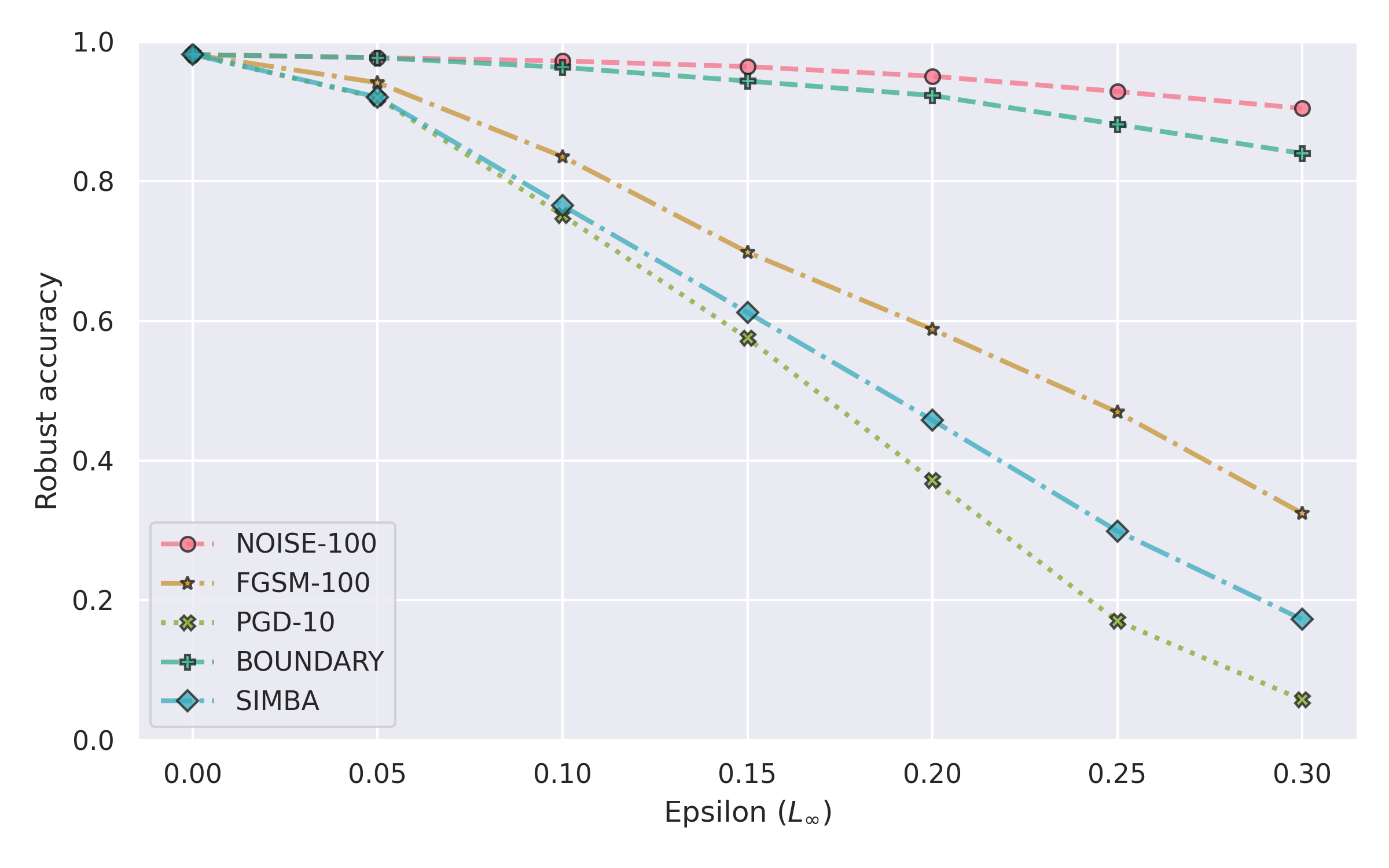}
        \caption{CT}
    \end{subfigure}
    ~
    \begin{subfigure}[b]{0.31\textwidth}
        \includegraphics[width=\textwidth]{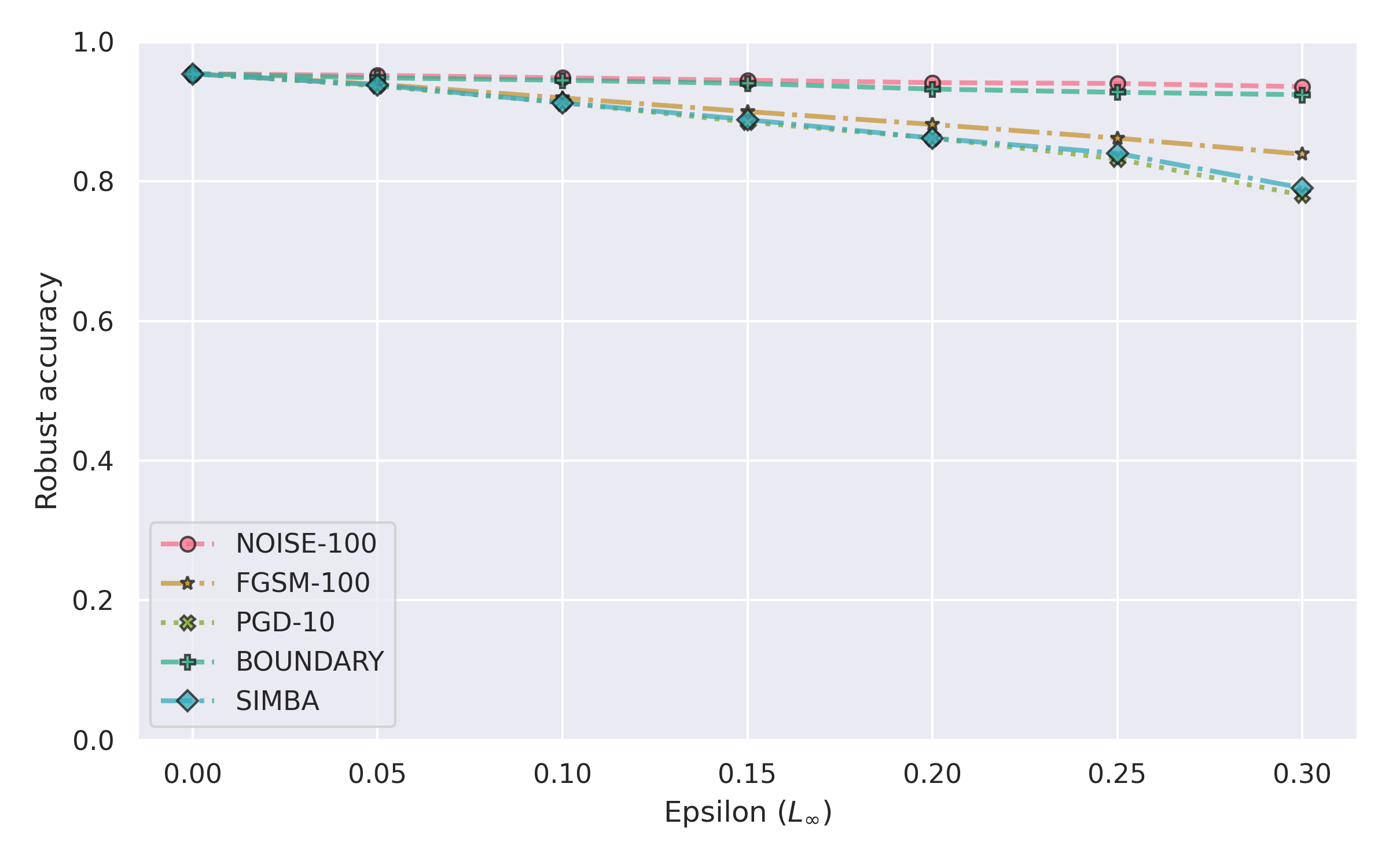}
        \caption{AT~\cite{madry2017towards}}
    \end{subfigure}
    ~
    \begin{subfigure}[b]{0.31\textwidth}
        \includegraphics[width=\textwidth]{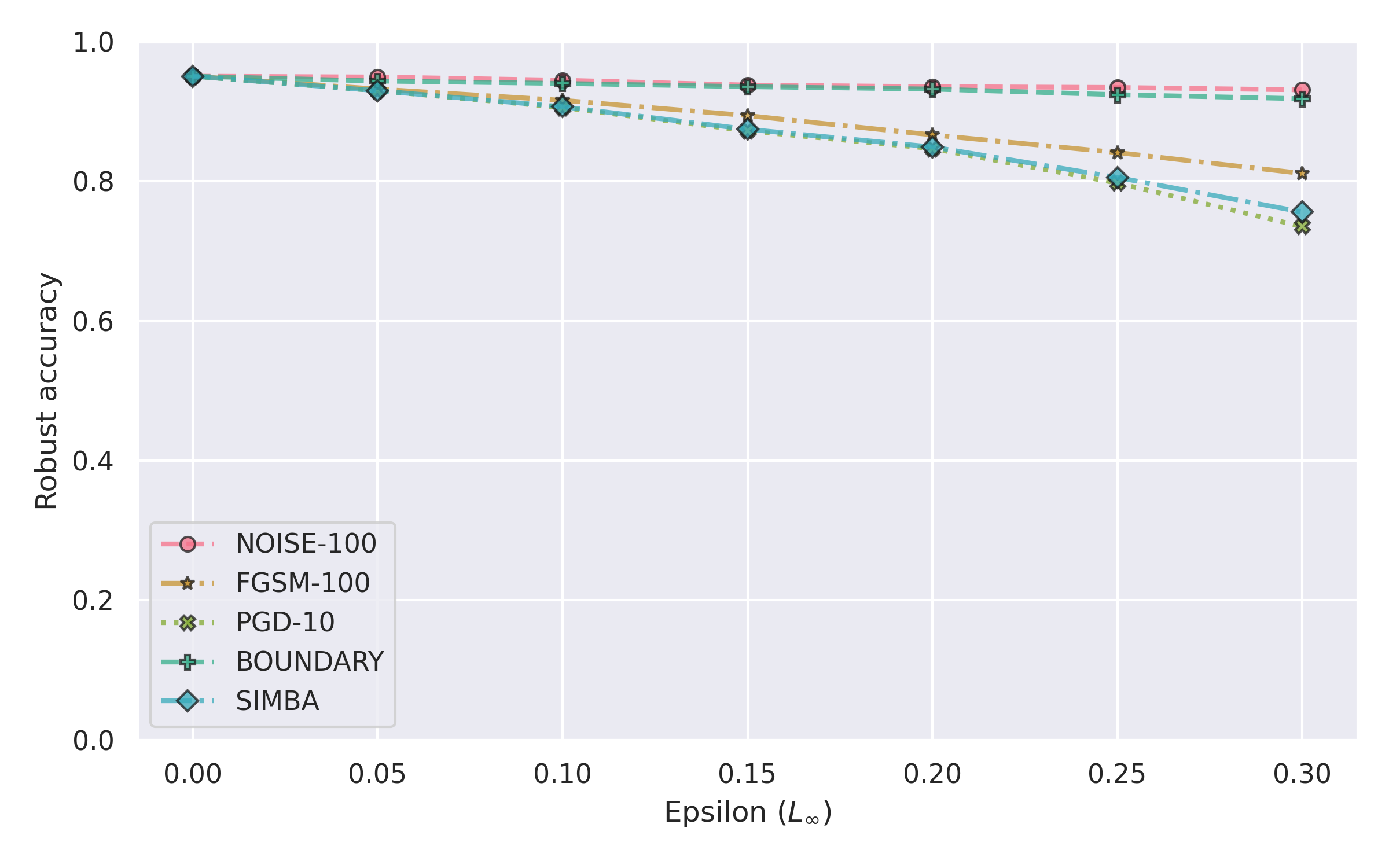}
        \caption{TR~\cite{zhang2019trades}}
    \end{subfigure}
    
    \begin{subfigure}[b]{0.31\textwidth}
        \includegraphics[width=\textwidth]{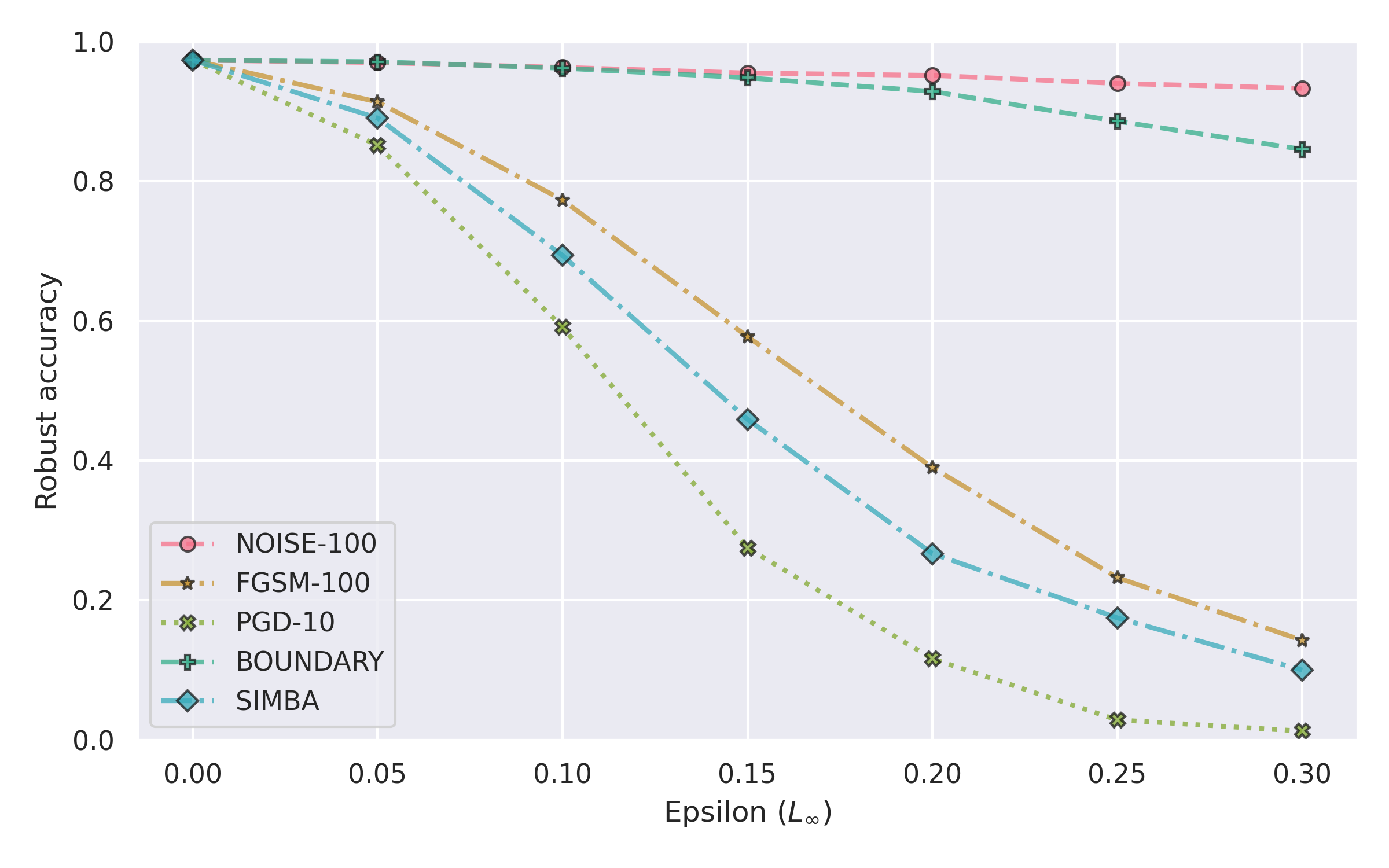}
        \caption{CT + GNLM~\cite{xie2019featuredenoising}}
    \end{subfigure}
    ~
    \begin{subfigure}[b]{0.31\textwidth}
        \includegraphics[width=\textwidth]{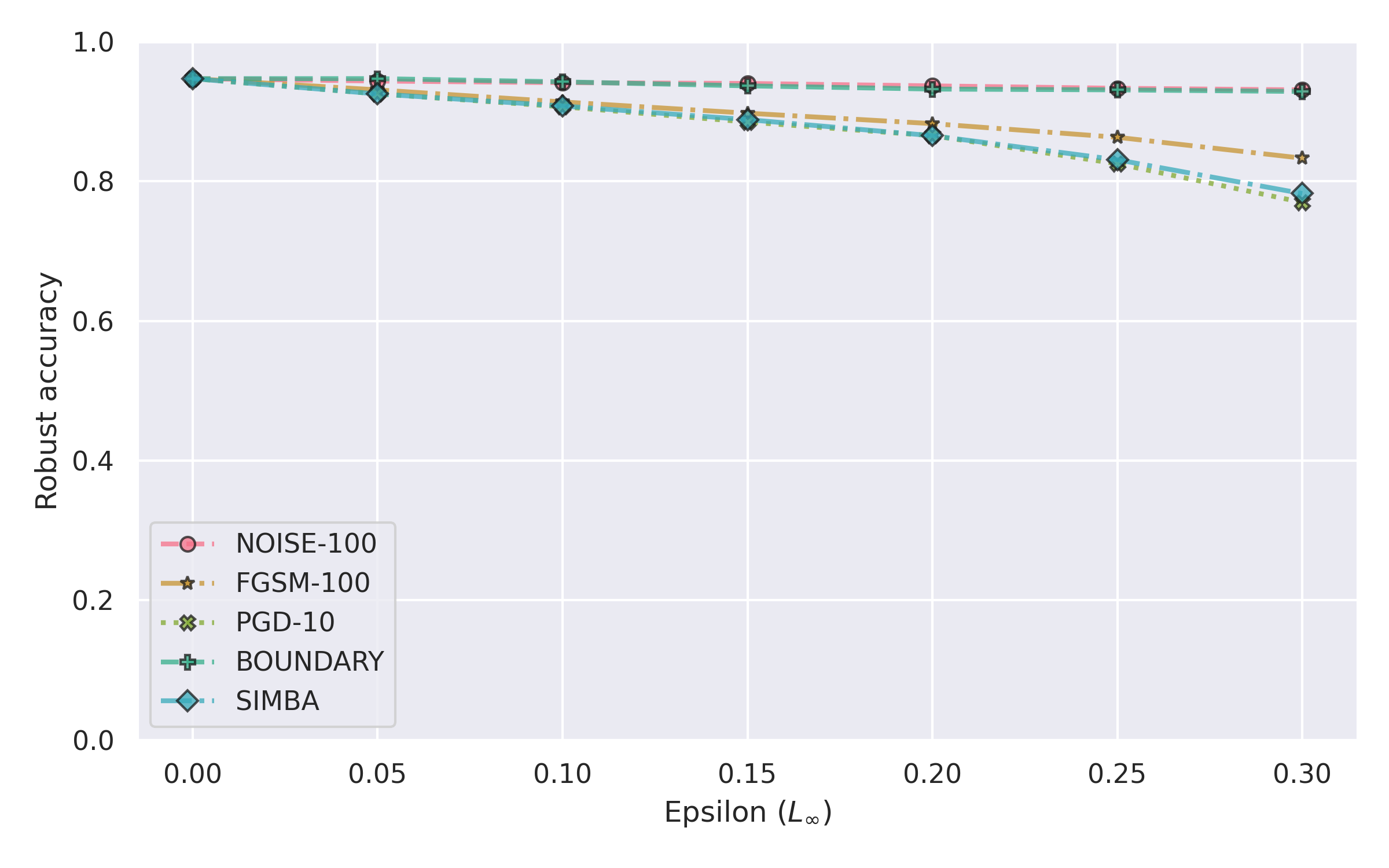}
        \caption{AT~\cite{madry2017towards} + GNLM~\cite{xie2019featuredenoising}}
    \end{subfigure}
    ~
    \begin{subfigure}[b]{0.31\textwidth}
        \includegraphics[width=\textwidth]{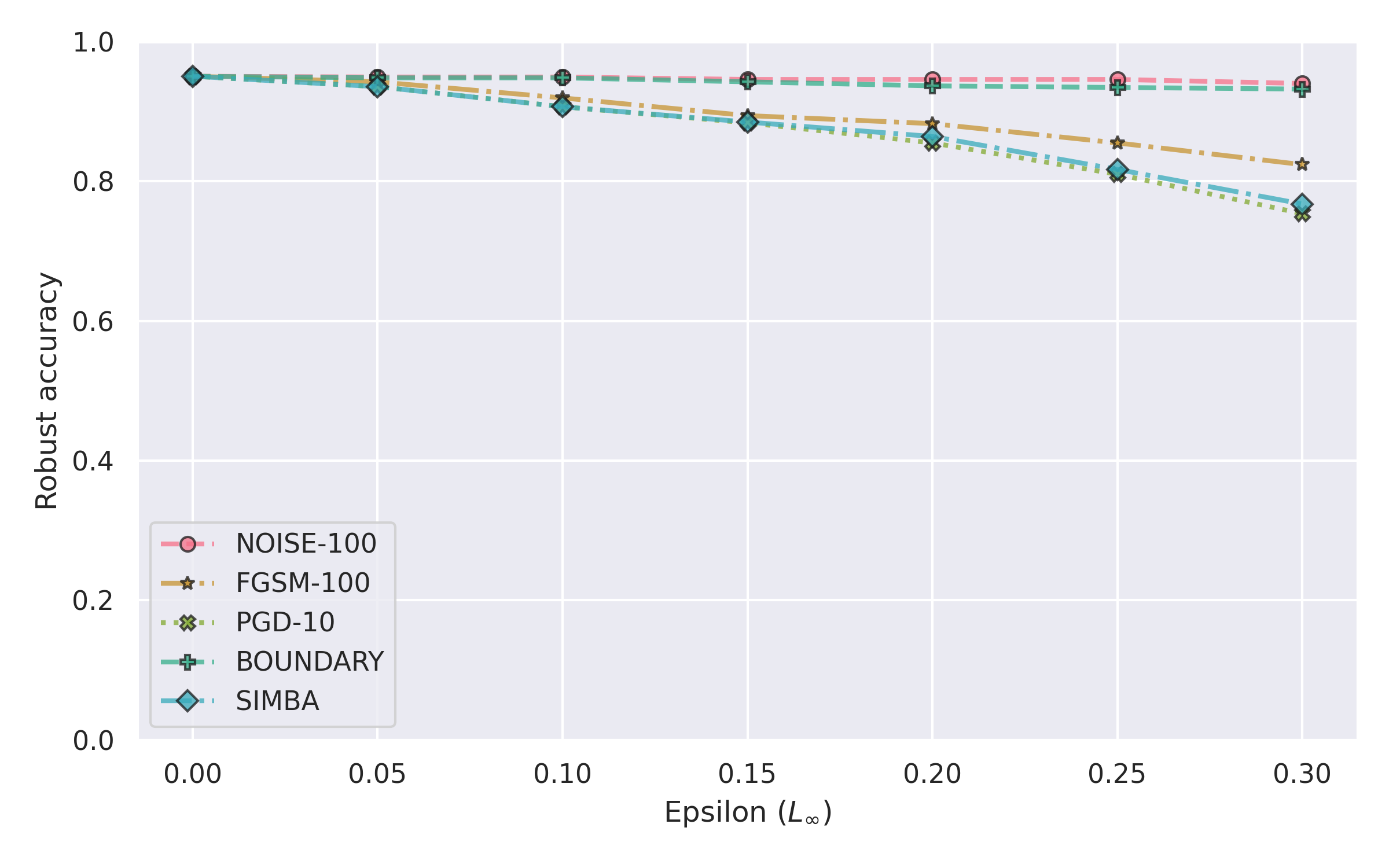}
        \caption{TR~\cite{zhang2019trades} + GNLM~\cite{xie2019featuredenoising}}
    \end{subfigure}

    \caption{Adversarial robustness curves computed for the different robust models using five different attacks. CT stands for conventional training, AT stands for adversarial training, TR stands for TRADES, and GNLM stands for Gaussian Non-Local Means.}
    \label{fig:main_results}
\end{figure}

Our main results are presented in Fig.~\ref{fig:main_results}. We report the robust accuracy of the classifier considering different attack methods on a range of different epsilon values starting from 0.05 to 0.3 with an increment of 0.05. These plots precisely capture the robustness of the model against adversarial attacks of different magnitudes. Since we trained a range of different TRADES models using different values of $1/\lambda$, we only report the best one out here, and explore the impact of these hyperparameters later in Section~\ref{sec:trades_hp}.
It is evident from the plot that conventional training results in poor robustness against these attacks, almost reducing the classifier's accuracy to 0\% when considering the worst-case adversary. However, when using defense methodologies such as adversarial training or TRADES, the model gains robustness against both white-box as well as black-box attacks with a slightly detrimental effect on the clean accuracy of the model.

\subsection{Quantifying the Impact of Denoising Operators}

\begin{figure}[t]
    \centering
    \begin{subfigure}[b]{0.31\textwidth}
        \includegraphics[width=\textwidth]{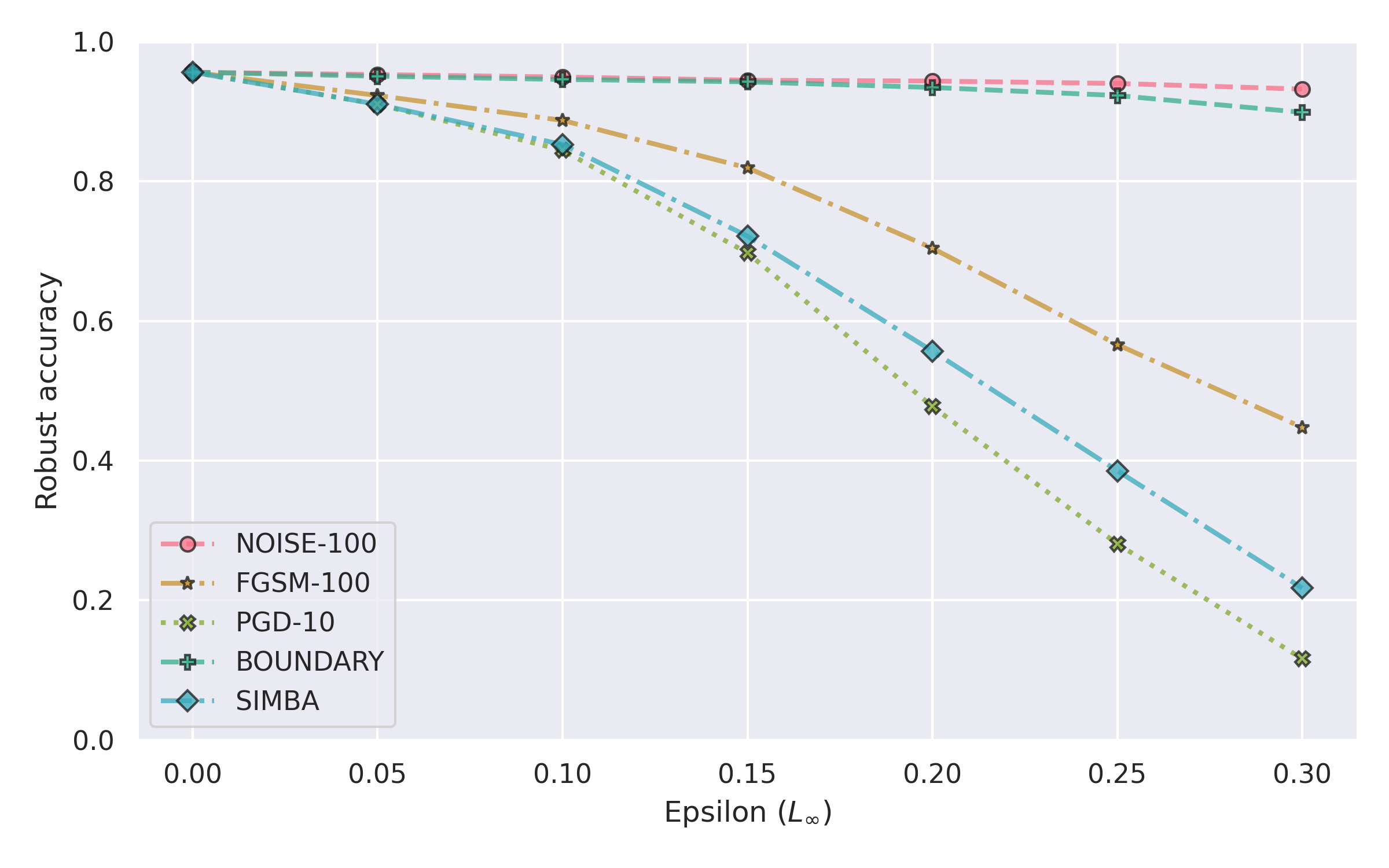}
        \caption{$1/\lambda=0.01$}
    \end{subfigure}
    ~
    \begin{subfigure}[b]{0.31\textwidth}
        \includegraphics[width=\textwidth]{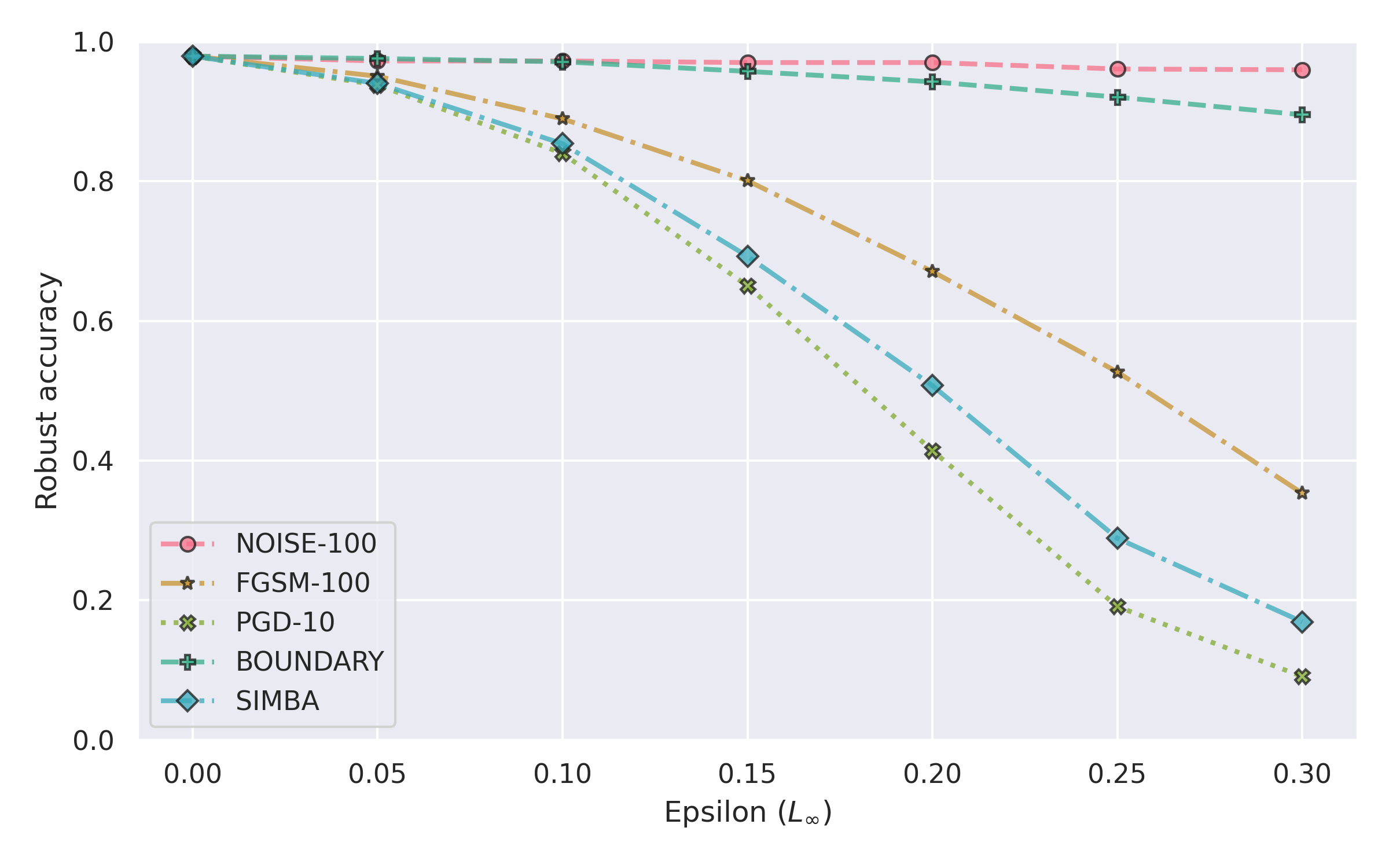}
        \caption{$1/\lambda=0.01$ + GNLM}
    \end{subfigure}
    ~
    \begin{subfigure}[b]{0.31\textwidth}
        \includegraphics[width=\textwidth]{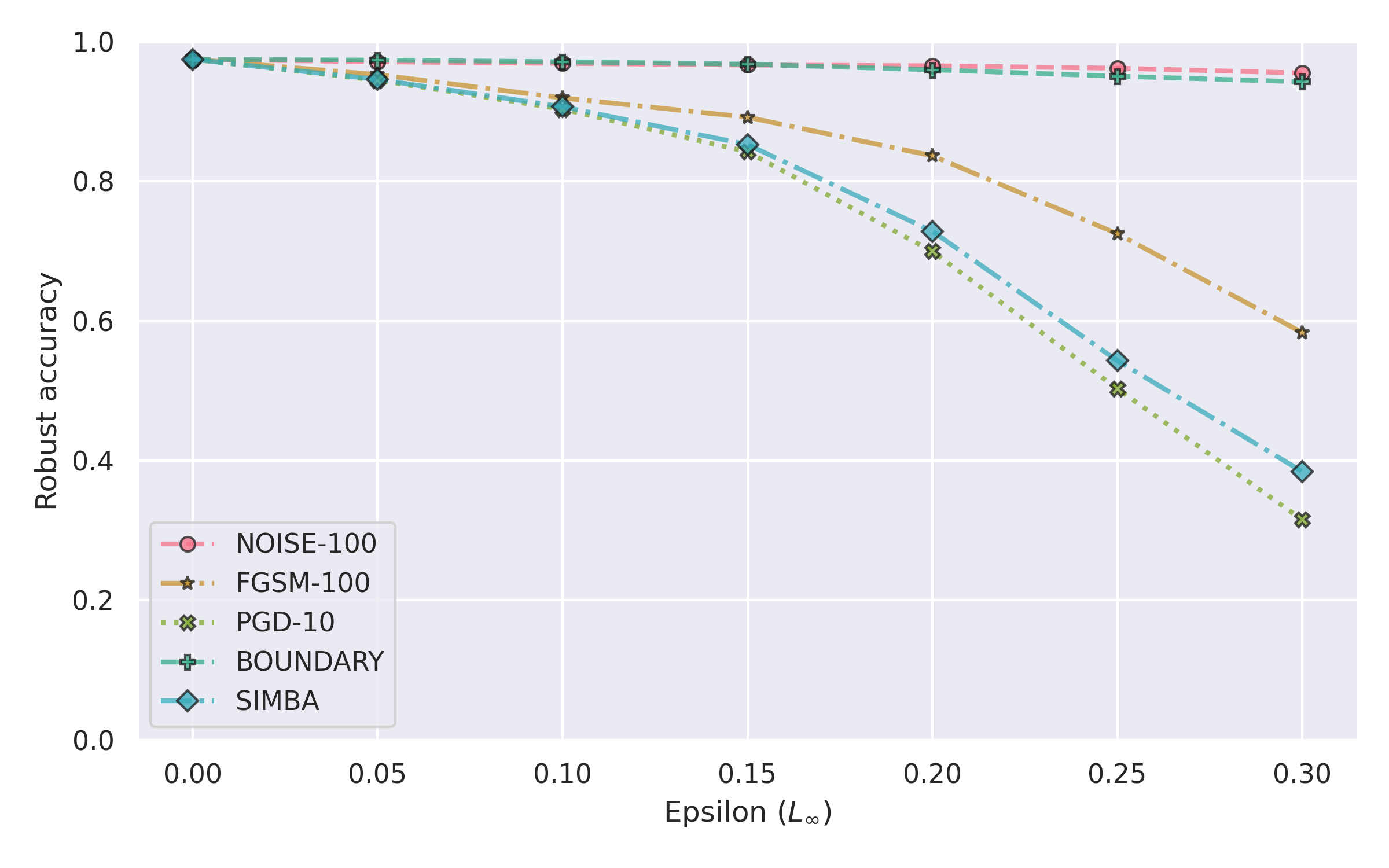}
        \caption{$1/\lambda=0.05$}
    \end{subfigure}
    
    \begin{subfigure}[b]{0.31\textwidth}
        \includegraphics[width=\textwidth]{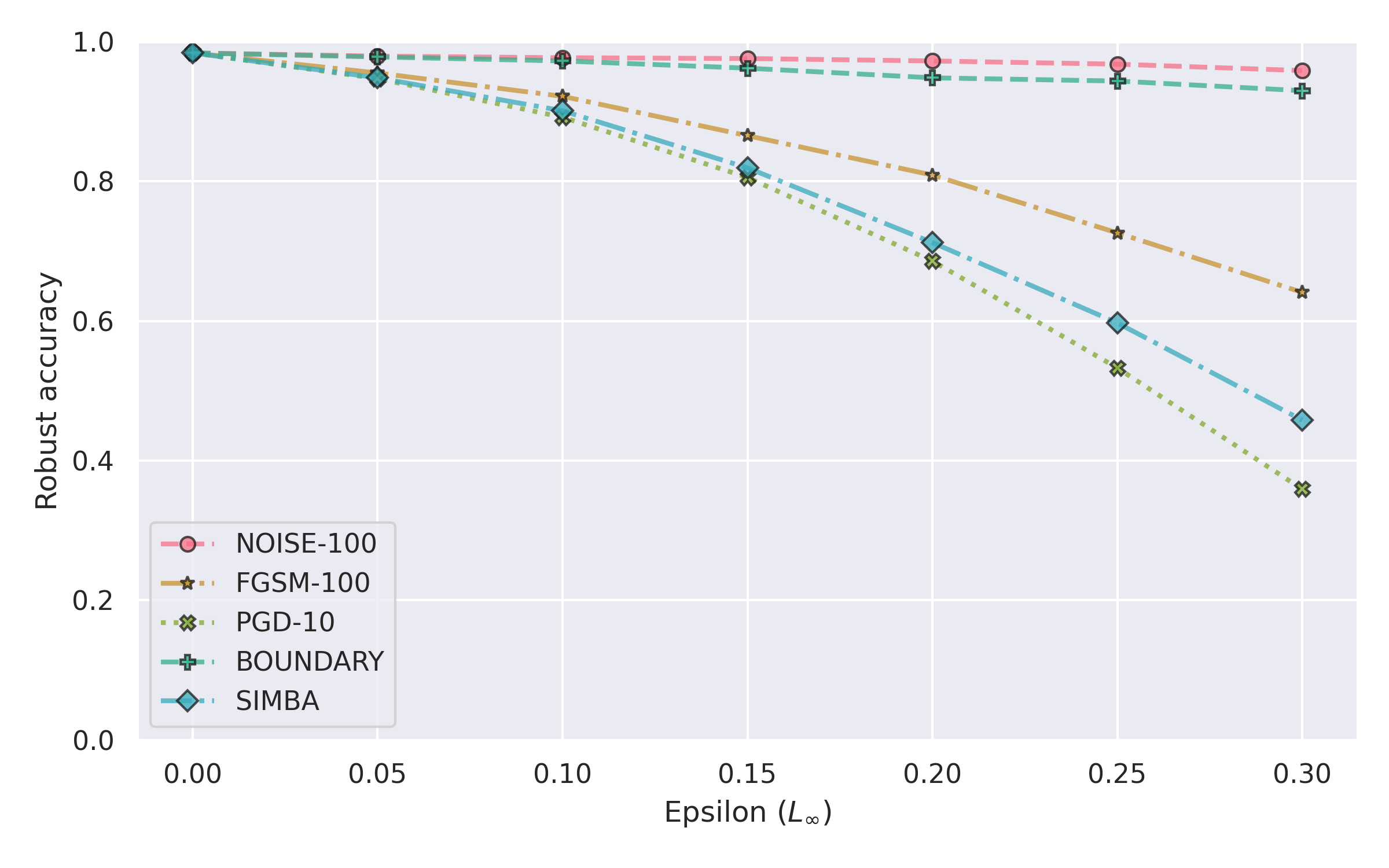}
        \caption{$1/\lambda=0.05$ + GNLM}
    \end{subfigure}
    ~
    \begin{subfigure}[b]{0.31\textwidth}
        \includegraphics[width=\textwidth]{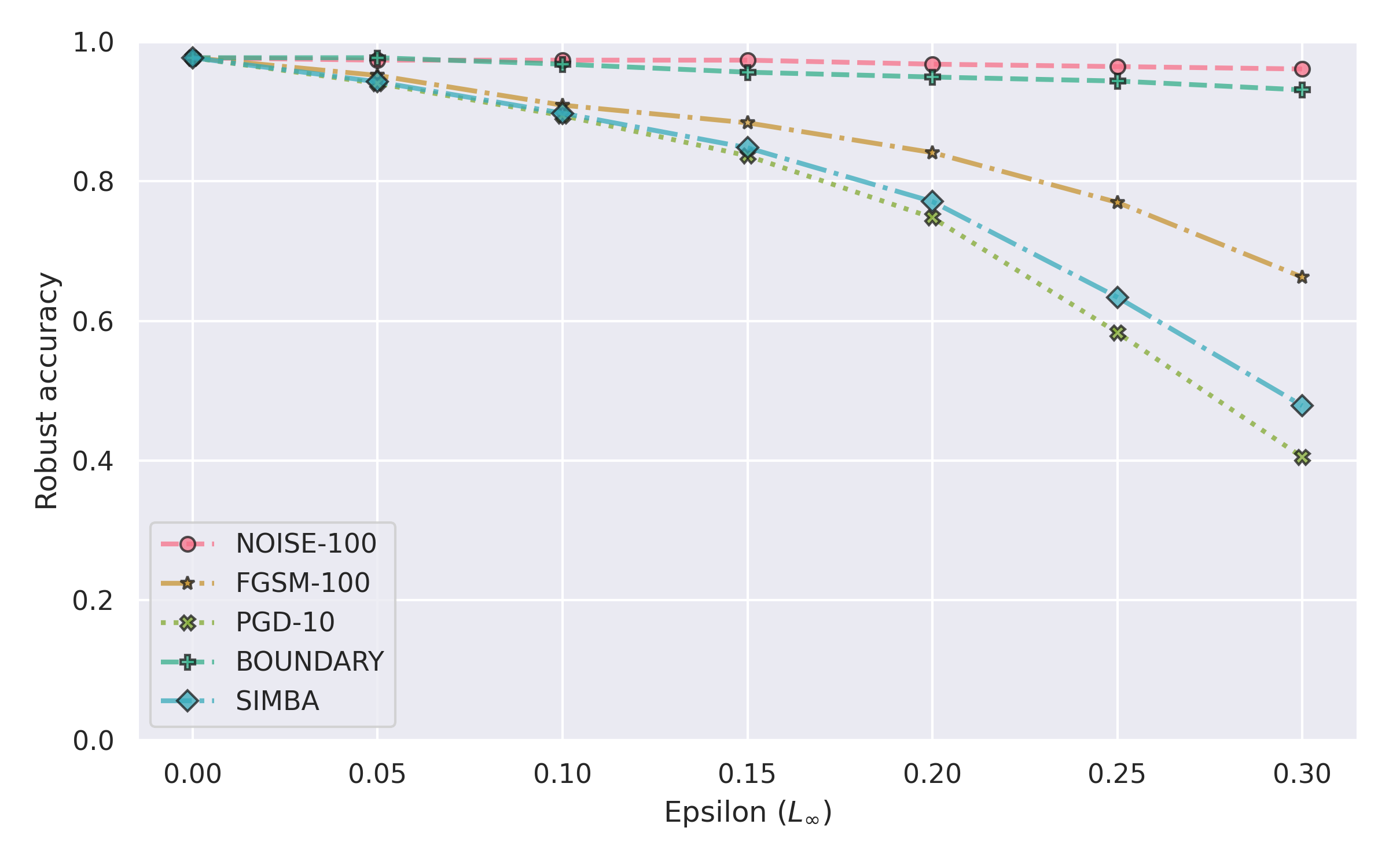}
        \caption{$1/\lambda=0.1$}
    \end{subfigure}
    ~
    \begin{subfigure}[b]{0.31\textwidth}
        \includegraphics[width=\textwidth]{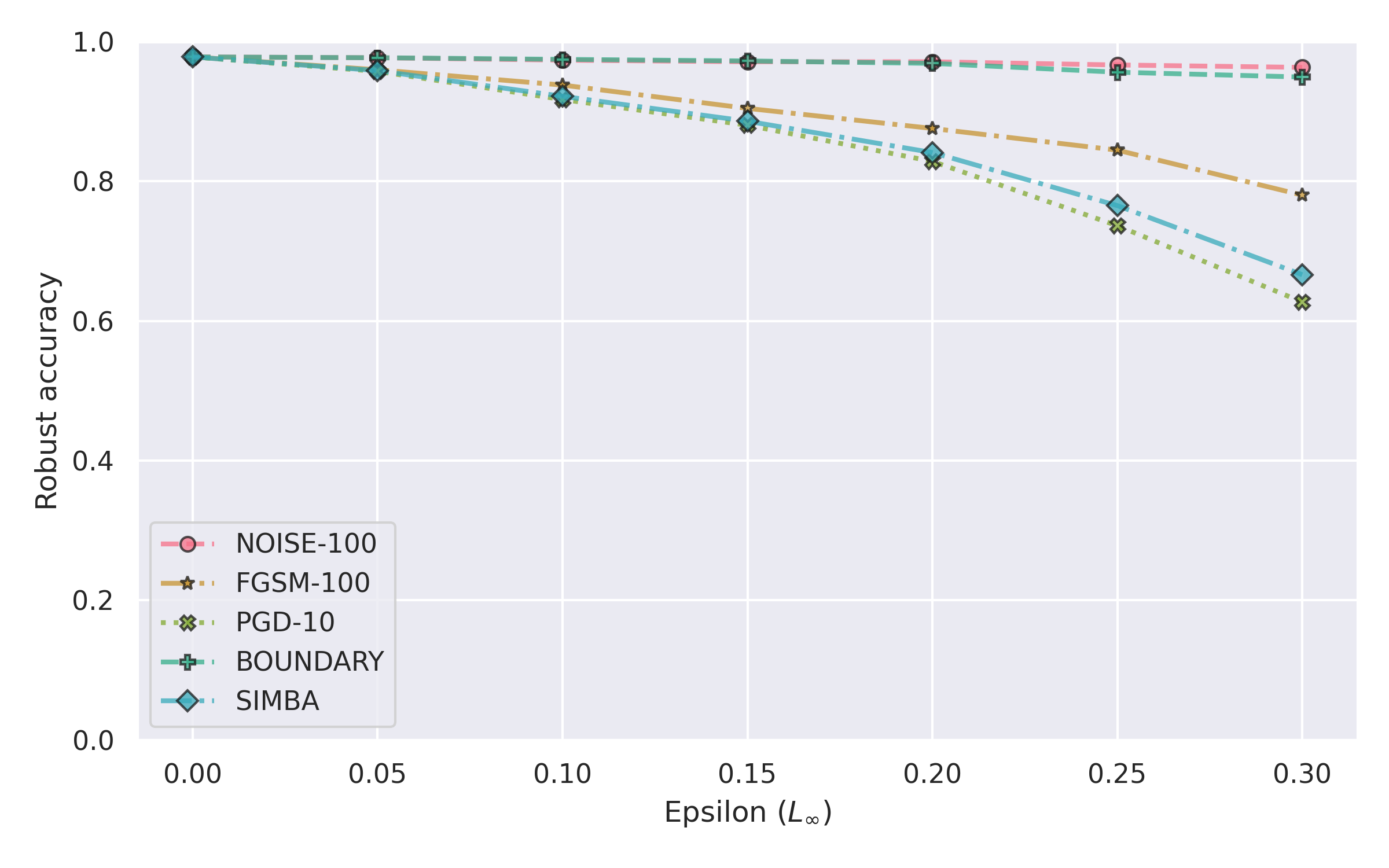}
        \caption{$1/\lambda=0.1$ + GNLM}
    \end{subfigure}
    
    \begin{subfigure}[b]{0.31\textwidth}
        \includegraphics[width=\textwidth]{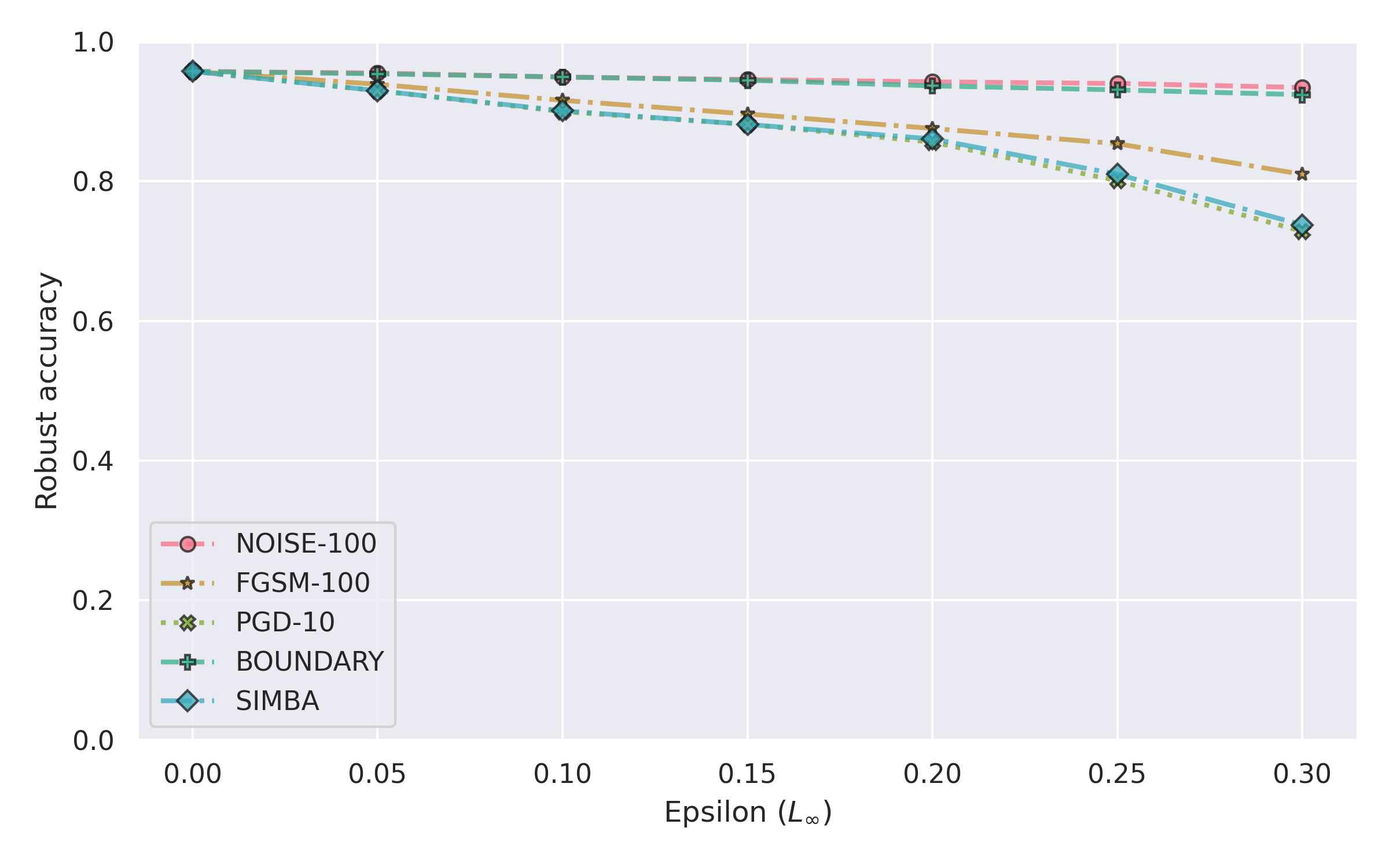}
        \caption{$1/\lambda=0.5$}
    \end{subfigure}
    ~
    \begin{subfigure}[b]{0.31\textwidth}
        \includegraphics[width=\textwidth]{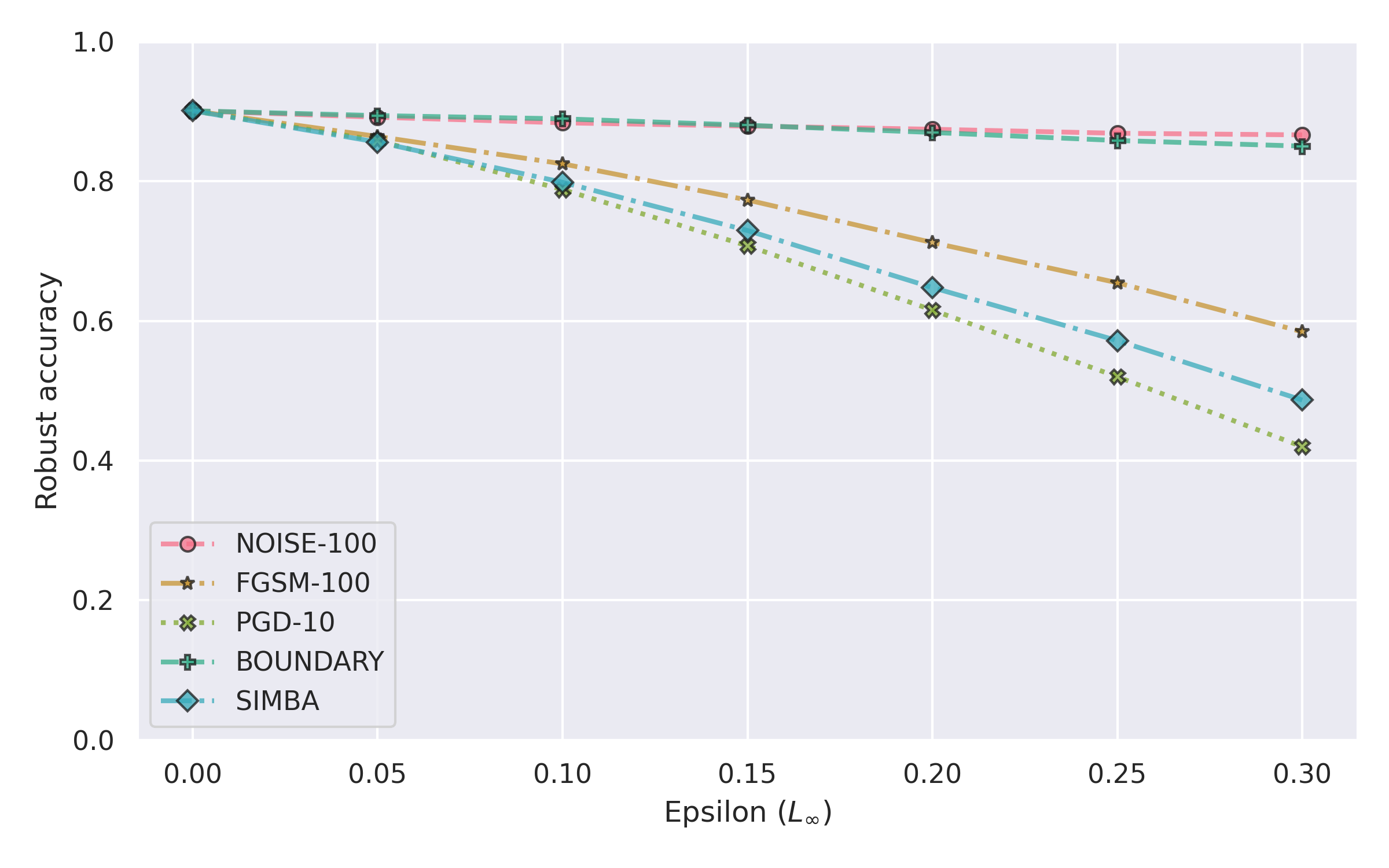}
        \caption{$1/\lambda=0.5$ + GNLM}
    \end{subfigure}
    ~
    \begin{subfigure}[b]{0.31\textwidth}
        \includegraphics[width=\textwidth]{Images/character_traj/robust_acc_trades_1_0.png}
        \caption{$1/\lambda=1.0$}
    \end{subfigure}
    
    \begin{subfigure}[b]{0.31\textwidth}
        \includegraphics[width=\textwidth]{Images/character_traj/robust_acc_trades_1_0_non_local_means-gaussian.png}
        \caption{$1/\lambda=1.0$ + GNLM}
    \end{subfigure}
    ~
    \begin{subfigure}[b]{0.31\textwidth}
        \includegraphics[width=\textwidth]{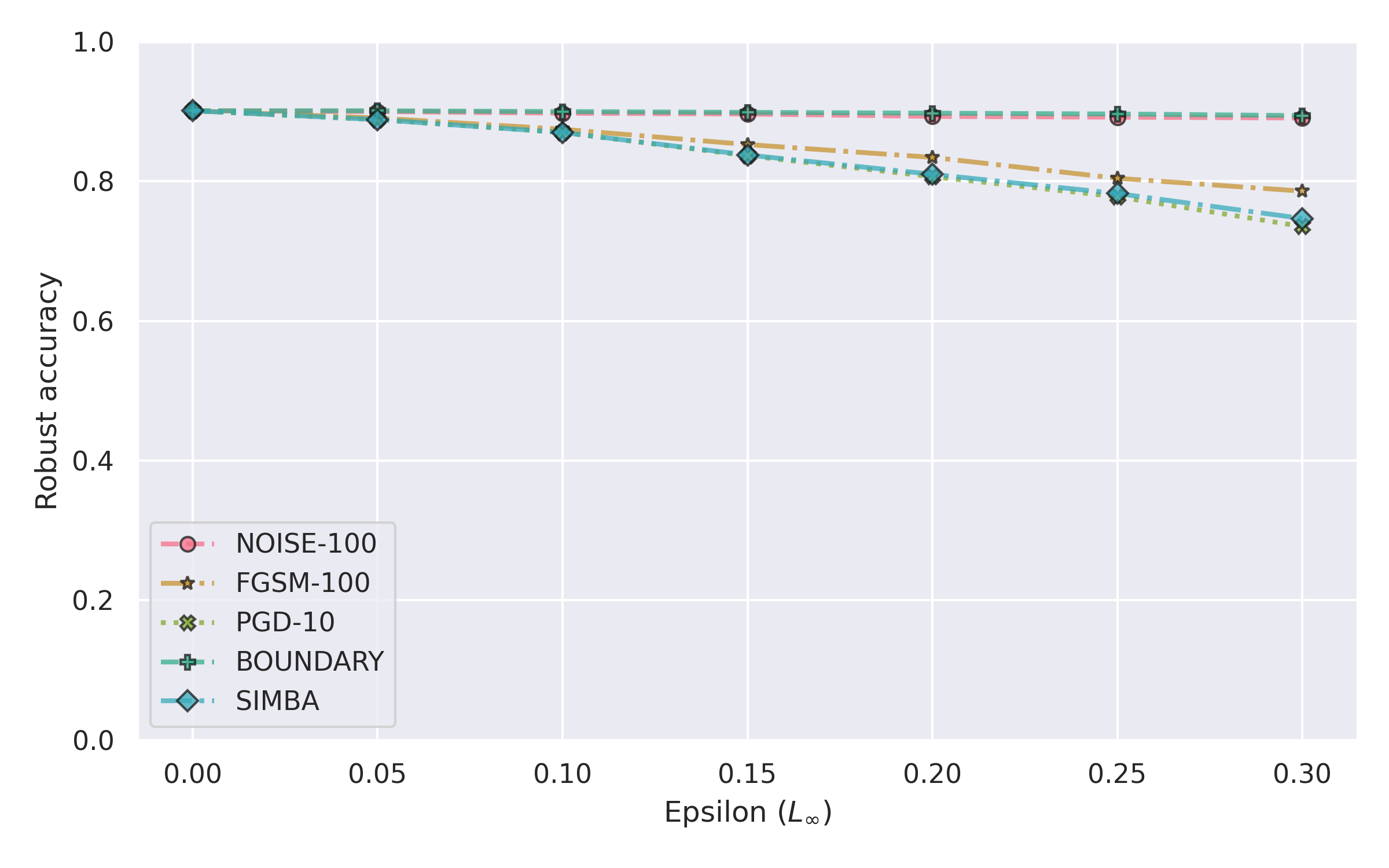}
        \caption{$1/\lambda=5.0$}
    \end{subfigure}
    ~
    \begin{subfigure}[b]{0.31\textwidth}
        \includegraphics[width=\textwidth]{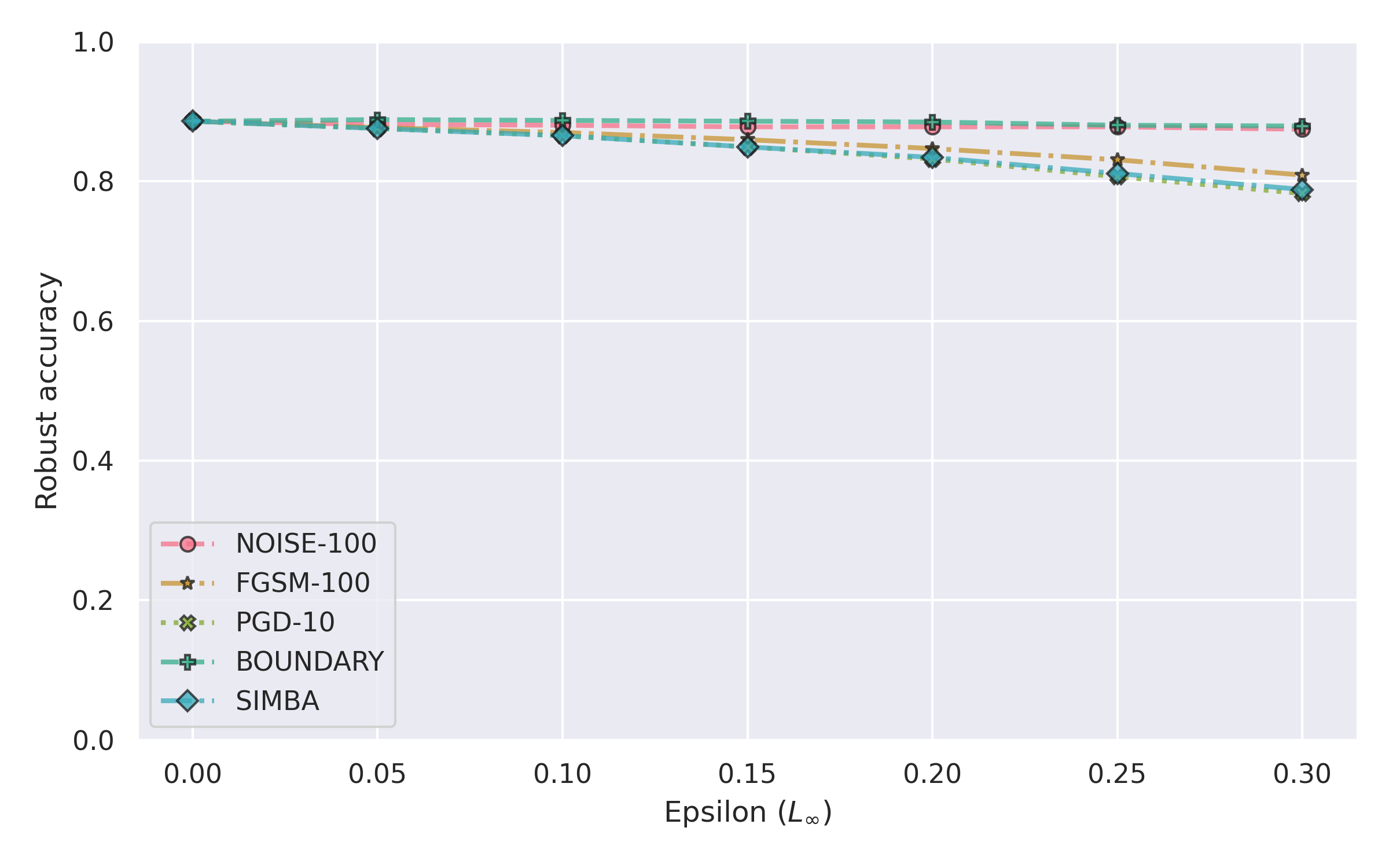}
        \caption{$1/\lambda=10.0$}
    \end{subfigure}
    
    \caption{Impact of hyperparameters on the robustness curves for TRADES.}
    \label{fig:trades_hp}
\end{figure}

\begin{table*}[t]
    \scriptsize
    \centering
    \caption{Clean accuracy of the different models}
    \begin{tabular}{c c c c c}
        \toprule
         \textbf{Defense} & \textbf{$1/\lambda$} & \textbf{Denoising Operator} & \textbf{Train Accuracy} & \textbf{Test Accuracy} \\
         \midrule
         - & - & - & 100.00\% & 98.16\% \\
         - & - & GNLM~\cite{xie2019featuredenoising} & 100.00\% & 97.35\% \\
         
         Adversarial Training~\cite{madry2017towards} & - & - & 99.13\% & 95.40\% \\
         Adversarial Training~\cite{madry2017towards} & - & GNLM~\cite{xie2019featuredenoising} & 98.70\% & 94.71\% \\
         
         TRADES~\cite{zhang2019trades} & 0.01 & - & 99.49\% & 95.63\% \\
         TRADES~\cite{zhang2019trades} & 0.01 & GNLM~\cite{xie2019featuredenoising} & 99.93\% & 97.93\% \\
         
         TRADES~\cite{zhang2019trades} & 0.05 & - & 99.71\% & 97.47\% \\
         TRADES~\cite{zhang2019trades} & 0.05 & GNLM~\cite{xie2019featuredenoising} & 99.57\% & 98.39\% \\
         
         TRADES~\cite{zhang2019trades} & 0.1 & - & 99.42\% & 97.70\% \\
         TRADES~\cite{zhang2019trades} & 0.1 & GNLM~\cite{xie2019featuredenoising} & 99.71\% & 97.81\% \\
         
         TRADES~\cite{zhang2019trades} & 0.5 & - & 99.13\% & 95.86\% \\
         TRADES~\cite{zhang2019trades} & 0.5 & GNLM~\cite{xie2019featuredenoising} & 89.08\% & 90.10\% \\
         
         TRADES~\cite{zhang2019trades} & 1.0 & - & 98.77\% & 95.05\% \\
         TRADES~\cite{zhang2019trades} & 1.0 & GNLM~\cite{xie2019featuredenoising} & 98.70\% & 95.05\% \\
         
         TRADES~\cite{zhang2019trades} & 5.0 & - & 96.02\% & 90.10\% \\
         TRADES~\cite{zhang2019trades} & 5.0 & GNLM~\cite{xie2019featuredenoising} & - & - \\
         
         TRADES~\cite{zhang2019trades} & 10.0 & - & 92.99\% & 88.61\% \\
         TRADES~\cite{zhang2019trades} & 10.0 & GNLM~\cite{xie2019featuredenoising} & 93.42\% & 87.46\% \\
         \bottomrule
    \end{tabular}
    \label{tab:clean_acc}
\end{table*}

Fig.~\ref{fig:main_results} also presents a comparison of using Gaussian Non-Local Means (GNLM) as a denoising operator alongside the use of different training schemes. Using denoising operator with conventional training results in inferior adversarial performance along with inferior clean accuracy (98.16\% vs 97.35\%) since the features are not optimized for this denoising operator. There is a slight drop in clean accuracy when switching from adversarial training to adversarial training with a denoising operator (95.40\% vs 94.71\%) alongside a minor drop in terms of robustness. This drop in robustness is primarily a consequence of the initial drop in clean accuracy as our evaluation metric is directly impacted by such changes. This drop is permissible for ImageNet classifiers where the accuracy even after adversarial training is only 35\%~\cite{xie2019featuredenoising}. However, for time-series datasets where the initial accuracy is already high, GNLM shows a detrimental effect on performance. 

In contrast, when using TRADES, the accuracy of the classifier remains the same with and without the denoising operator. The denoising operator also positively impacts the robustness of the model. In comparison to adversarial training, there is no impact on clean accuracy when using TRADES with $1/\lambda=1.0$. Table~\ref{tab:clean_acc} summarizes the clean accuracies of the model under different settings for a direct comparison.



\subsection{Sensitivity to Regularization Hyperparameters} \label{sec:trades_hp}

Fig.~\ref{fig:trades_hp} visualizes the robustness curves for different values of the hyperparameter $1/\lambda$ used when training the robust model using TRADES. We also list the clean accuracies of the models in Table~\ref{tab:clean_acc} for a direct comparison. It is evident from the table and the figure that higher regularization leads to lower clean accuracy as expected alongside higher robustness against adversarial attacks. It is important to note that the network failed to converge in many cases when using $1/\lambda > 1.0$ and GNLM denoising operator.

\subsection{Attacked Examples}

\begin{figure}[t]
    \centering
    
    \begin{subfigure}[b]{0.31\textwidth}
        \includegraphics[width=\textwidth]{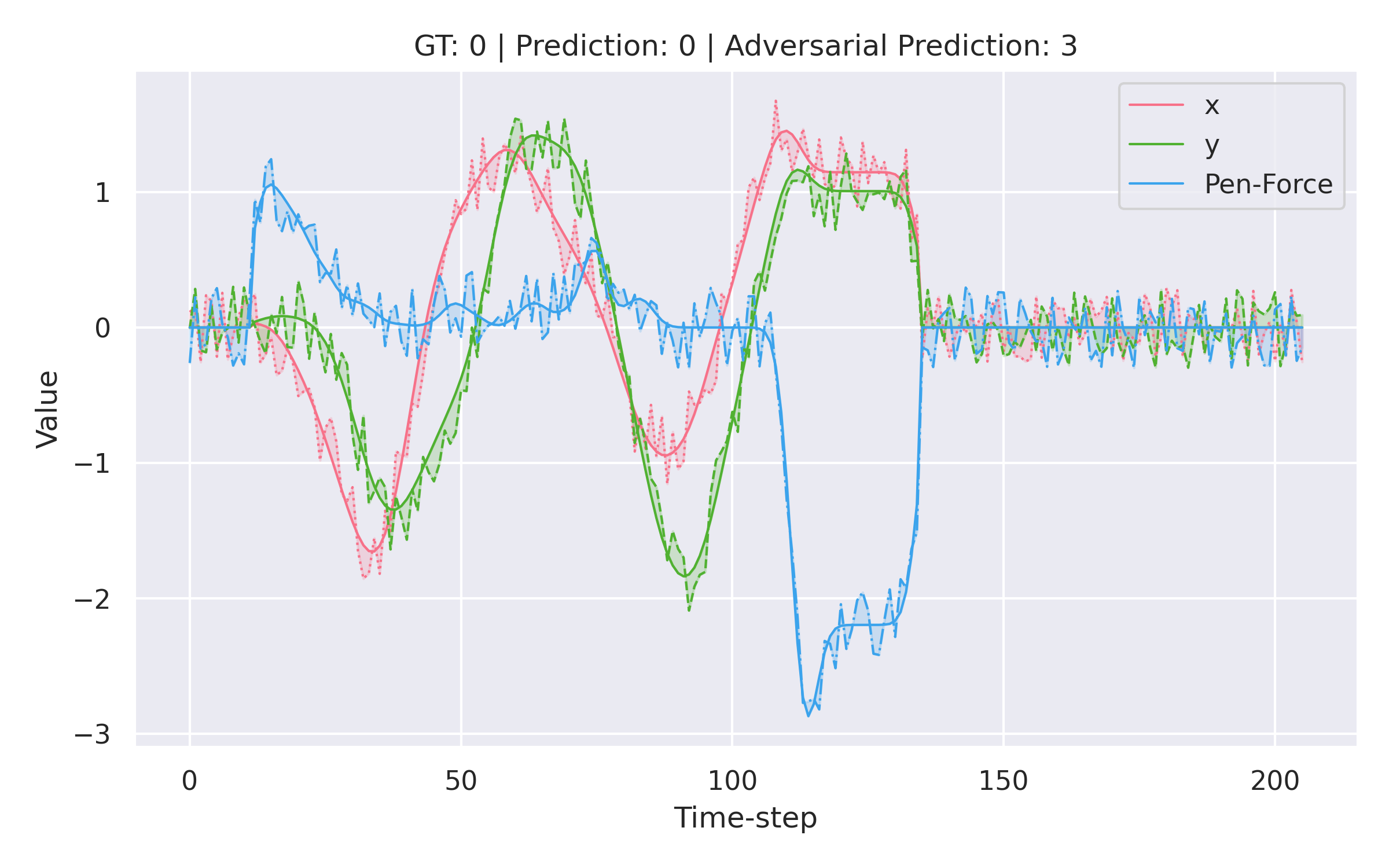}
        \caption{NOISE-100}
    \end{subfigure}
    ~
    \begin{subfigure}[b]{0.31\textwidth}
        \includegraphics[width=\textwidth]{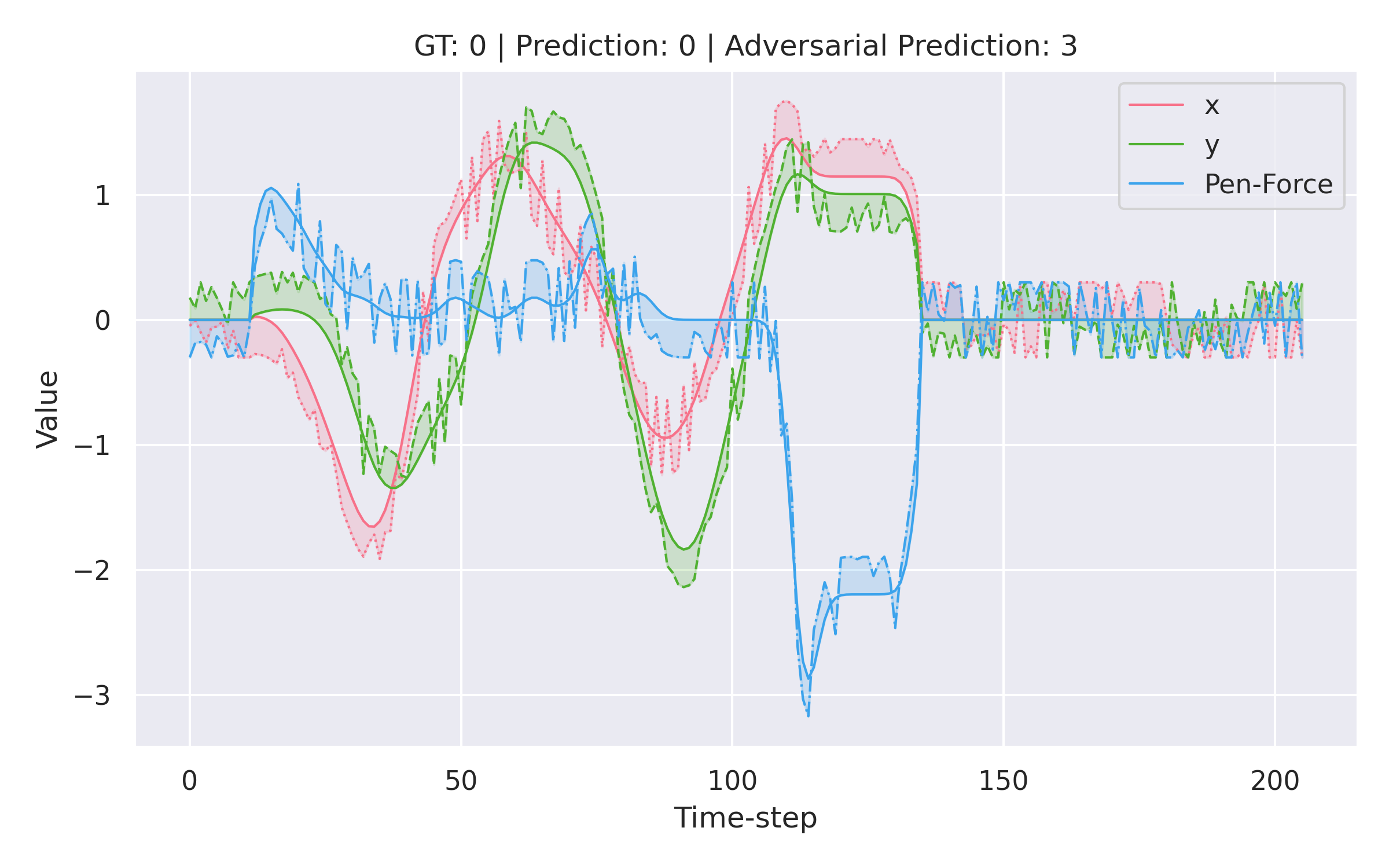}
        \caption{FGSM-100}
    \end{subfigure}
    ~
    \begin{subfigure}[b]{0.31\textwidth}
        \includegraphics[width=\textwidth]{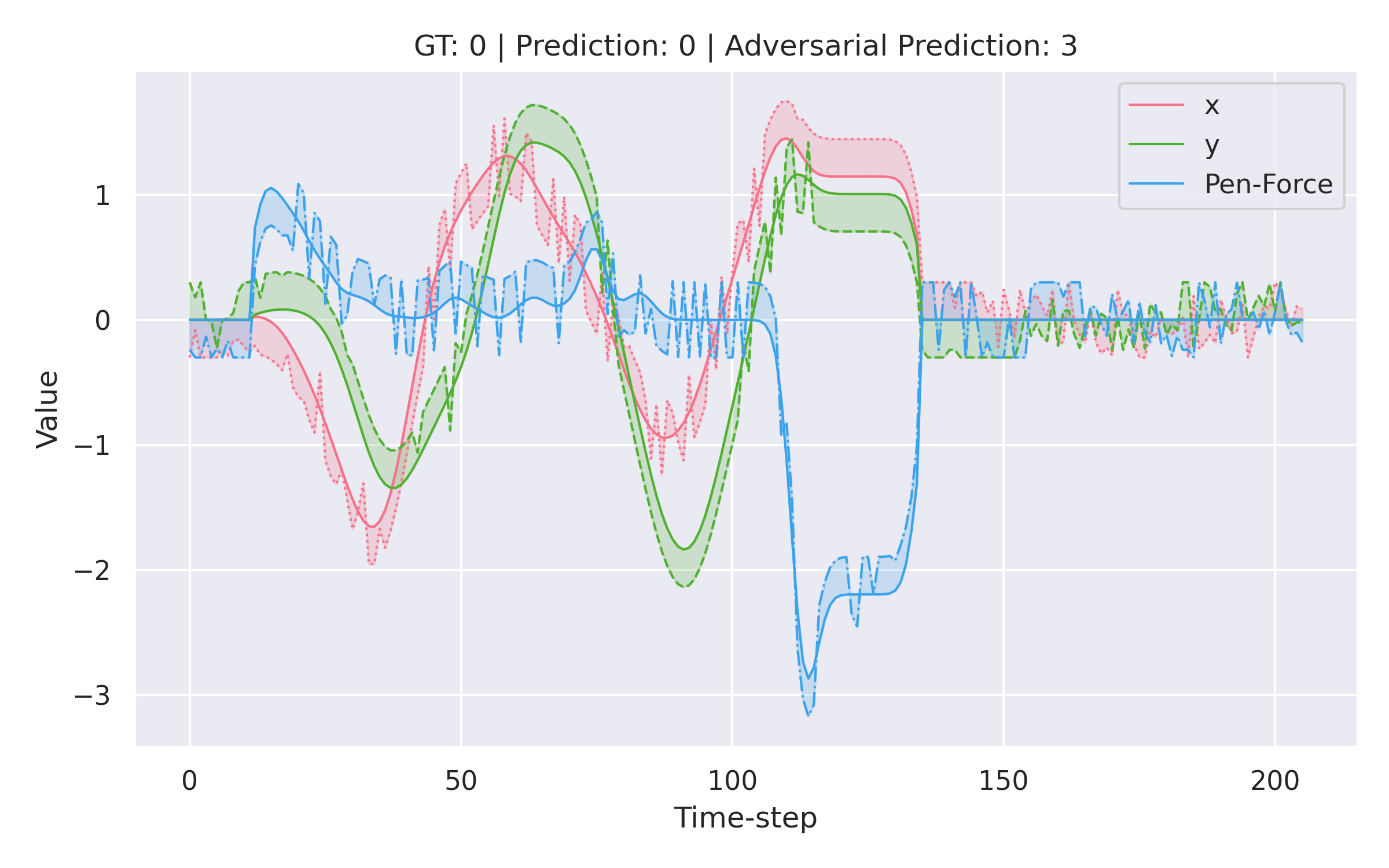}
        \caption{PGD-10}
    \end{subfigure}
    
    \begin{subfigure}[b]{0.31\textwidth}
        \includegraphics[width=\textwidth]{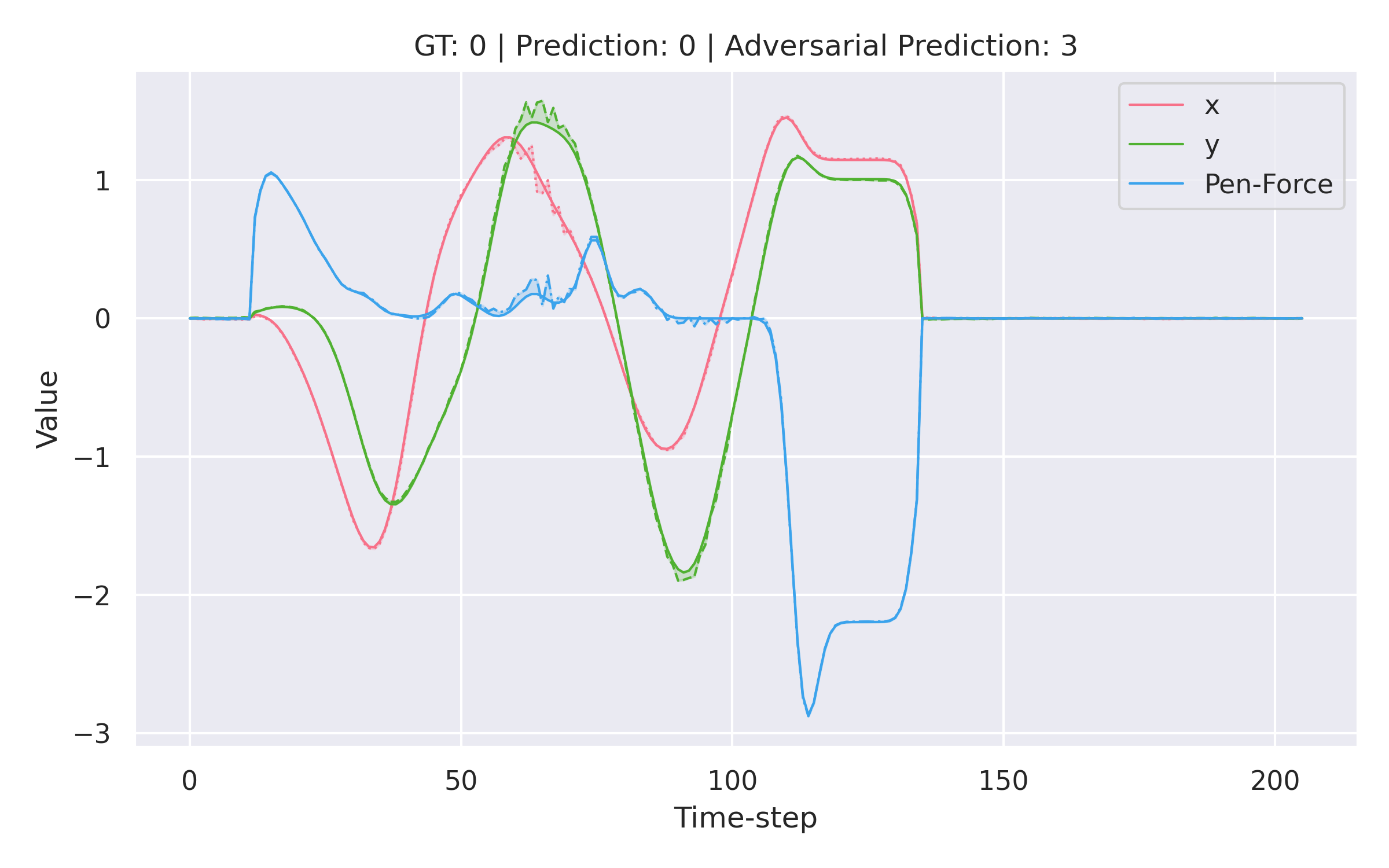}
        \caption{Boundary~\cite{brendel2017decision}}
    \end{subfigure}
    ~
    \begin{subfigure}[b]{0.31\textwidth}
        \includegraphics[width=\textwidth]{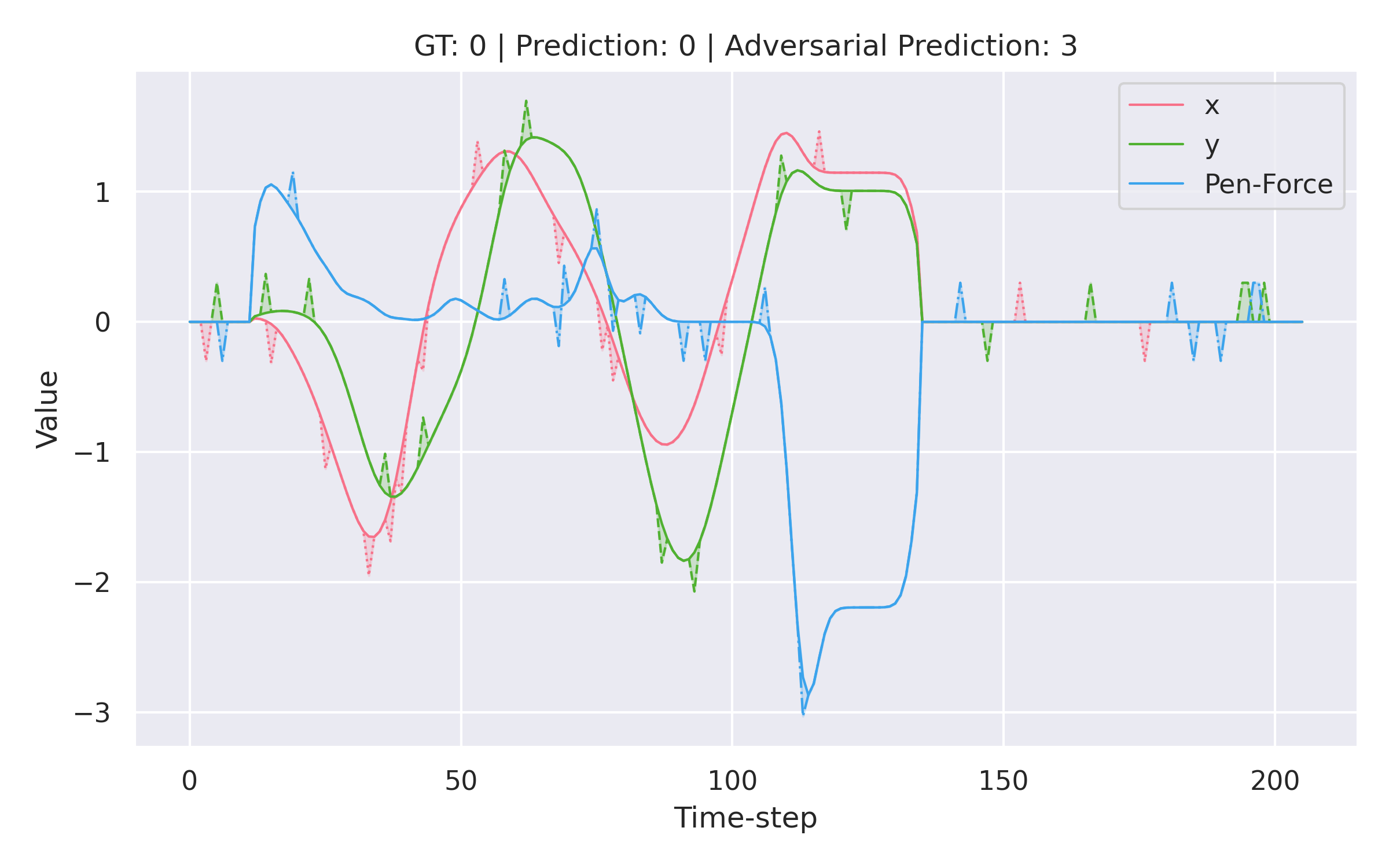}
        \caption{SIMBA~\cite{simba}}
    \end{subfigure}
    
    \caption{Generated adversarial examples using an $\epsilon$ of 0.3. The original signal is highlight using a solid line while the attacked signal is represented using different line styles. The shaded area highlights the difference between the two signals.}
    \label{fig:adv_ex}
\end{figure}

    
    

Fig.~\ref{fig:adv_ex} presents a particular example from the character trajectories dataset on the undefended model. 
We visualized examples generated from a rather high value of epsilon i.e. 0.3. This is to ensure that the differences between different attacks are properly highlighted.
It is interesting to note that all attacks changed the label to the same target class 3 indicating that the two classes are similar in the feature space.
Almost all attacks exhausted the $L_{\infty}$ perturbation budget except for boundary and SIMBA attack as boundary attack minimizes the $L_2$ norm of the perturbation while SIMBA additionally minimizes the $L_0$ norm of the perturbation alongside the $L_{\infty}$ norm.

\section{Conclusion} \label{sec:conclusion}

This paper establishes an important benchmark regarding the robustness of time-series classification models trained using different adversarial defense techniques. Our analysis shows that the defenses evaluated for visual modality provide similar robustness against adversarial attacks on time-series data.

Future work should be mainly targeted towards the evaluation of these adversarial attacks for regression networks. While it is easy to quantify the impact in terms of success rate for classification networks, this is much harder to report when considering real-valued outputs. Another important direction is to compare provable robustness methods on time-series data and evaluate their efficacy as compared to the defenses considered here.

%
%
\bibliographystyle{splncs04}
\bibliography{ecml.bib}

\end{document}